# BanglaSummEval: Reference-Free Factual Consistency Evaluation for Bangla Summarization

**Ahmed Rafid[1,*], Rumman Adib[1,*], Fariya Ahmed[1,*]**

**Ajwad Abrar[1], Mohammed Saidul Islam[2]**

[1]Department of Computer Science and Engineering,
Islamic University of Technology, Bangladesh

[2]York University, Canada

`{ahmedrafid, rummanadib, fariyaahmed, ajwadabrar}@iut-dhaka.edu`,
`saidulis@yorku.ca`

## Abstract

Evaluating factual consistency is essential for reliable text summarization, particularly in high-stakes domains such as healthcare and news. However, most existing evaluation metrics overlook Bangla, a widely spoken yet under-resourced language, and often depend on reference summaries. We introduce BanglaSummEval, a reference-free, question-answering-based framework for evaluating factual consistency in Bangla summarization. The proposed method assesses both factual accuracy and content coverage through automatically generated questions and answers derived from the source document and the summary. A single multilingual instruction-tuned language model handles question generation, question answering, candidate answer extraction, and question importance weighting. This unified design reduces system complexity and computational cost. To capture semantic consistency beyond surface-level overlap, we use BERTScore-Recall for answer comparison. We validate BanglaSummEval on 300 human-written summaries from educational and medical domains, demonstrating strong correlation with expert human judgments (Pearson's $r = 0.694$, Spearman's $\rho = 0.763$). By providing interpretable, step-wise diagnostics alongside reliable evaluation scores, BanglaSummEval offers a practical and transparent solution for factual consistency evaluation in low-resource language settings.

## 1 Introduction

The integration of Large Language Models (LLMs) into healthcare and other service sectors is increasing rapidly. For example, there is a growing use of LLMs for medical report summarization (Zhang et al., 2025). Professionals often have to deal with large volumes of data, where relevant information is frequently buried under irrelevant statements. LLMs can assist by summarizing reports and filtering out important information (Khan et al., 2023). Therefore, it is essential that LLM-generated summaries retain important information while remaining factually consistent with the source text.

Consequently, an evaluation metric that can precisely measure factual consistency between a source document and its summary is crucial for comparing LLM performance in terms of generating summaries. We explored several factual evaluation metrics and found that QuestEval (Scialom et al., 2021) provides a robust evaluation framework without requiring human-annotated reference summaries. QuestEval addresses key limitations of traditional metrics through its reference-free, question-answering-based approach, evaluating both factual consistency and content relevance.

However, none of the existing evaluation metrics support Bangla, the seventh most spoken language in the world, with over 240 million speakers. While traditional lexical overlap metrics such as ROUGE and BLEU are widely used for summary evaluation, they are fundamentally limited for factual consistency assessment, as they rely on surface-level similarity rather than semantic accuracy. These limitations are particularly severe for morphologically rich languages like Bangla, where paraphrasing and flexible word order are common. A factual consistency evaluation metric for Bangla would facilitate the development of more effective Bangla summarization models. In this paper, we present `BanglaSummEval`, an adaptation of the QuestEval framework for

---

[*]These authors contributed equally to this work.

Bangla. Beyond providing Bangla language support, our adaptation improves the transparency of LLM evaluation by enabling stepwise analysis of the evaluation process. This allows researchers to explicitly identify where LLMs fail and pinpoint instances of factual inconsistency or hallucination. Such diagnostic capability facilitates targeted model improvements rather than treating factual evaluation as an end-to-end score. Our major contributions are as follows:

1. We introduce `BanglaSummEval`, the first factual consistency evaluation metric that supports Bangla.

2. We evaluate `BanglaSummEval`, demonstrating its strong correlation with human judgments on factual consistency.

3. We enhance the transparency of LLM-based evaluation by incorporating stepwise analysis and explicit error localization.This enables precise identification of factual failure points.

Figure [1] illustrates the architecture of `BanglaSummEval`.

## 2 Related Work

Domains like healthcare, education, news are highly sensitive to factual errors, where reliability is the priority. Some of the key influential metrics currently available for such critical evaluation are QAGS (Tam et al., 2023), SummaC (Laban et al., 2022), FactScore (Min et al., 2023), QuestEval (Scialom et al., 2021) , QAFactEval (Fabbri et al., 2022) which illustrate the core approaches currently in this field.

QuestEval is an advanced metric that evaluates factuality in summaries and text generation by generating and answering questions using both the generated output and the source. Unlike many metrics, QuestEval does not need gold-reference annotations and assesses both recall and precision by generating questions from each side. It delivers robust correlation with human assessments and facilitates explainable diagnostics, highlighting gaps in specific information coverage.

SummaC decomposes a machine-generated summary into individual factual claims or sentences, then formulates the natural language inference (NLI) (Bowman et al., 2015) model to predict if the source supports, refutes, or is neutral towards each claim. FactScore extracts atomic factual statements from generated outputs and validates them against the source using retrieval and inference. QAFactEval is a question-answering-centric metric which creates a large pool of QA pairs derived from the candidate text and answers them using the source, then compares the answers with reference-ground-truth answers or using confidence scores.

None of the aforementioned highly cited metrics have out-of-the-box support for Bangla. In fact there are no established factual consistency metrics designed specifically for Bangla, nor major open-source tools or benchmarks tailored to this language. This lack of coverage presents a significant research opportunity considering the growing use of LLMs and the distinct linguistic characteristics in Bangla.

Our proposed metric `BanglaSummEval` works to fill this gap using an approach similar to QuestEval (QA- based and referenceless). Prior studies have shown QuestEval's superior correlation with human judgment in factual consistency assessments compared to metrics like BERTScore and ROUGE. Thus our metric does not require human-written reference summaries or labels, making it more scalable for low-resource languages like Bangla. It produces interpretable, question-based outputs, facilitating deeper analysis of model errors.

## 3 Methodology

In this section, we describe our adaptation of the QuestEval framework for evaluating Bangla summarization. While we retain the core mathematical formulation of precision and recall defined in QuestEval, our method, `BanglaSummEval` (see Figure [1]), introduces a unified model architecture with model specific answer confidence calculation and semantic scoring to suit low-resource Bangla evaluation.Though we largely retain the architecture and core pipeline of QuestEval, our major contributions lie in the extensions required to make the framework effective and reliable for Bangla.This includes identifying pre-trained and fine-tuned models suitable for this architecture in Bangla and evaluating their performance using task-specific datasets. `BanglaSummEval`, instantiates four key components: question answering (QA), question generation (QG), candidate answer extraction (NER) and the query weighter. It uses a single unified large language model (LLM). This unified approach reduces deployment complexity and computational overhead.

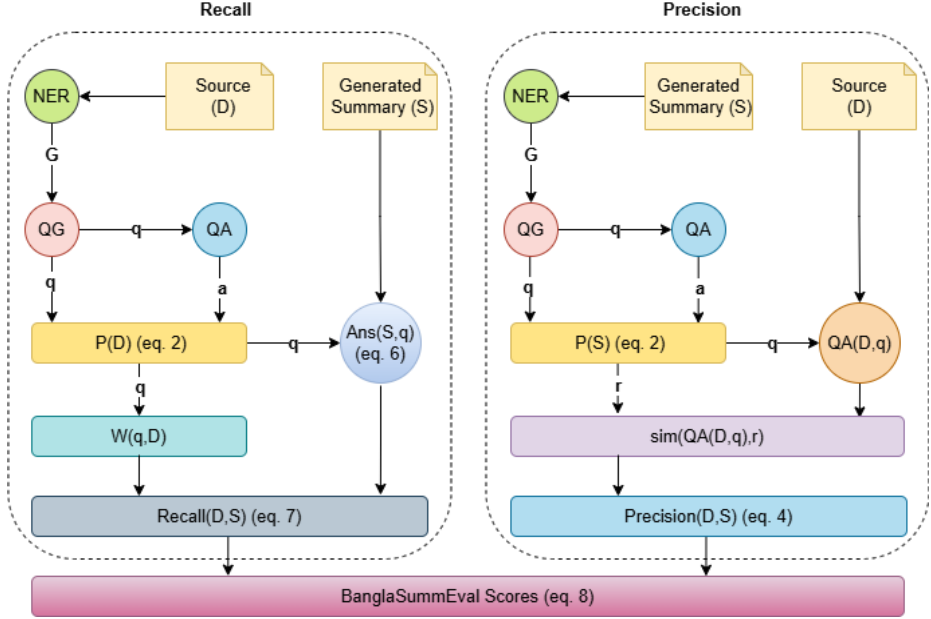

Figure 1: Overview of the proposed `BanglaSummEval` architecture.

### 3.1 Unified Model Architecture

We employ a 4-bit quantized version of the Qwen3-14B-Instruct model (Yang et al., 2025) provided by Unsloth named Qwen3-14B-bnb-4bit as the backbone for all four components (QA, QG, NER and query weighter).

For selecting the best backbone, a systematic benchmarking was conducted on the four large language models: BanglaT5, mT5-Base, Qwen3-8B-bnb-4bit, and Qwen3-14B-bnb-4bit. These models have been evaluated on the BanglaRQA (Ekram et al., 2022) corpus. We evaluated each model's question answering (QA) and question generation (QG) capabilities using 150 passages from the dataset. For QA evaluation, we prompted each model to answer two ground-truth questions per passage and measured answer similarity using BERTScore (Precision/Recall/F1) and semantic similarity. For QG evaluation, we prompted each model to generate two questions per passage and compared the generated questions against ground-truth questions using the same metrics. This yielded 300 QA pairs and 300 QG pairs for comprehensive assessment of both capabilities.

| Semantic similarity | BERT-P | BERT-R | BERT-F1 |
|---|---|---|---|
| **BanglaT5** | | | |
| QA | 0.55 | 0.55 | 0.65 | 0.59 |
| QG | 0.61 | 0.61 | 0.68 | 0.64 |
| **mT5-Base** | | | |
| QA | 0.59 | 0.59 | 0.64 | 0.61 |
| QG | 0.61 | 0.61 | 0.64 | 0.62 |
| **Qwen3-8B-bnb-4bit** | | | |
| QA | 0.62 | 0.62 | **0.73** | 0.67 |
| QG | 0.75 | 0.75 | 0.79 | 0.77 |
| **Qwen3-14B-bnb-4bit** | | | |
| QA | **0.63** | **0.63** | 0.72 | **0.67** |
| QG | **0.76** | **0.76** | **0.82** | **0.79** |

Table 1: Evaluation Metrics Comparison for BanglaT5, mT5-Base, Qwen3-8B-bnb-4bit, and Qwen3-14B-bnb-4bit on QA and QG Tasks (BanglaRQA Dataset)

As shown in Table 1, the model with the best performance for all evaluation metrics (BERTScore-F1: 0.67 QA, 0.79 QG) was Qwen3-14B-bnb-4bit. Qwen3-8B-bnb-4bit was a close competitor to Qwen3-14B-bnb-4bit, and the performance of BanglaT5 and mT5 was quite low which makes the effectiveness of instruction tuning in multiple languages highly evident. Qwen3-14B-bnb-4bit was therefore chosen based on excellent multi-lingual ability in low-resource environments, the ability to follow the Bangla instruction set and resource efficiency. Through our single model paradigm based on task-specific statements shown in Appendix A.1, it is ensured that our entire evaluation framework is lightweight and can be executed on a low-resource environment such as a free tier cloud GPU environment (Kaggle/Colab).

### 3.2 Candidate Answer Extraction

Reference-less evaluation requires extracting information or facts from the text to serve as ground-truth answers for generating questions. Following previous work (Wang et al., 2020), we focus on named entities and nouns as candidate answers.

Large language models have been shown to perform comparably to supervised baselines on NER tasks, with particularly strong performance in low-resource settings (Dai et al., 2023), and Qwen3's extensive multilingual pre-training across 119 languages (Yang et al., 2025) makes it well-suited for Bengali named entity extraction. Thus, to keep our framework light, we reused Qwen3-14B-bnb-4bit as a Bangla named entity and noun extractor using specific prompts (Appendix A.1.3), which aligns with our unified model approach and removes the necessity of using any external NER models which would consume additional resources in an already resource constrained setup.

The NER component takes a context $C$ and produces a set of unique candidate answers $\mathcal{G}(C)$, consisting of named entities and nouns present in $C$.

### 3.3 Constructing Answer-Conditioned Question Sets

Both precision and recall metrics in our framework rely on a set of high-quality question-answer pairs derived from a source text. For a given context $C$ (either the source document $D$ or the summary $S$), we construct a filtered set of pairs $\mathcal{P}(C)$ through a similarity based filtering process. First, for each candidate answer $r \in \mathcal{G}(C)$, we generate a corresponding question $q$ using the QG prompt:

$$q = \text{QG}(C, r) \quad (1)$$

To ensure the generated question is valid and answerable we feed the generated question $q$ back into the QA component with context $C$ to obtain a predicted answer $a = \text{QA}(C, q)$. The pair $(q, r)$ is accepted if and only if the semantic similarity between the predicted answer $a$ and the original candidate $r$ is equal to or greater than a threshold $\tau$ (see Appendix A.3):

$$(q, r) \in \mathcal{P}(C) \iff \text{sim}(a, r) \geq \tau \quad (2)$$

where $\text{sim}(\cdot, \cdot)$ is a semantic similarity function described below.

### 3.4 Semantic Similarity Metric

Unlike the original QuestEval, which uses token-level F1 score for answer comparison, we employ BERTScore Recall (Zhang et al., 2020) to capture semantic similarity in Bangla. Semantic similarity is used only for comparison between ground truths and answer generated on questions asked since document-level similarity metrics are fundamentally limited for factuality evaluation. As Ye et al. observe, "similarity-based metrics are insufficient for precisely detecting factual errors, because high similarity cannot guarantee factual consistency" (Ye et al., 2024). By decomposing evaluation into atomic QA pairs, we overcome this limitation by isolating specific factual claims for targeted comparison.

The similarity function is defined as:

$$\text{sim}(x, y) = \text{BERTScore}_R(y, x) \quad (3)$$

where $x$ is the reference answer and $y$ is the candidate answer.

For the document-level summary evaluation, the traditional similarity metrics such as ROUGE or BLEU underperform since they are based on surface-level lexical overlap, which cannot effectively handle the morphological richness of the Bangla language or paraphrasing. Contrastively, BERTScore is appropriate for answer-level comparison in our QA framework. The key distinction lies in granularity: our decomposition into atomic

| Metric | Pearson_r | Pearson_p | MAE | RMSE |
|---|---|---|---|---|
| *Semantic Metrics* | | | | |
| BERTScore-Recall | **0.673** | **6.12E-41** | 29.317 | 43.194 |
| BERTScore-F1 | 0.624 | 1.02E-33 | 29.896 | 42.975 |
| Cosine Similarity | 0.592 | 1.04E-29 | **26.511** | **33.845** |
| *Lexical Metrics* | | | | |
| chrF | 0.666 | 8.88E-40 | 27.273 | 37.215 |
| Token-F1 | 0.496 | 4.61E-20 | 41.343 | 55.217 |
| BLEU | 0.249 | 1.30E-05 | 60.761 | 72.391 |
| Exact Match | 0.296 | 1.91E-07 | 54.433 | 69.831 |
| *Character-level Metrics* | | | | |
| CER Similarity | 0.470 | 7.43E-18 | 39.744 | 54.495 |
| WER Similarity | 0.366 | 5.69E-11 | 50.266 | 65.489 |

Table 2: Correlation of automatic metrics with human judgments on Bangla.

QA pairs transforms the problem from holistic document similarity to targeted verification of individual factual claims. For answer-level comparison, the answer is typically a short entity mention or noun phrase, for which BERTScore's semantic embeddings excel at capturing meaning despite morphological variation between different expressions of the same fact. Unlike document-level metrics that conflate multiple quality dimensions, answer-level BERTScore focuses exclusively on the semantic equivalence of specific facts, which directly addresses factual consistency verification. By utilizing contextualized embeddings from multilingual models such as XLM-RoBERTa-base (Conneau et al., 2019), BERTScore-Recall can robustly capture semantic meaning across lexically different but factually equivalent expressions, making it highly suitable for factual verification in our decomposed abstractive summarization evaluation framework.

To empirically validate this choice, we benchmarked lexical, semantic, and hybrid metrics against human-annotated scores using 300 stratified samples from the BanglaRQA dataset (Ekram et al., 2022). Answers were generated using TigerLLM-1B-it (Raihan and Zampieri, 2025) and evaluated by human annotators (the annotation guideline is provided in Appendix C), each with over 14 years of experience in the Bangla language. Annotators rated factual accuracy on a continuous scale from 0.0 to 1.0. Correlation with human judgments was measured using Pearson's correlation coefficient ($r$) and Spearman's rank correlation coefficient ($\rho$), along with MAE, RMSE, and L2-norm deviation.

As shown in Table 2, semantic metrics exhibit stronger correlation with human judgments than lexical and character-level alternatives. BERTScore-Recall, computed using XLM-RoBERTa-base, achieves the highest Pearson correlation ($r = 0.673$) and the lowest $p$-value, indicating a strong and statistically significant alignment with human judgments. This is followed by BERTScore-F1 ($r = 0.624$) and cosine similarity ($r = 0.592$), both showing moderate-to-strong correlations. These results confirm that semantic similarity metrics are more robust for evaluating factual consistency in Bangla, as they capture meaning beyond surface-level lexical overlap. Consequently, we select BERTScore-Recall for `BanglaSummEval`, as it enables accurate comparison even when answers are phrased differently but convey equivalent meaning.

### 3.5 Precision: Verification against Source

Precision measures the extent to which information in the summary $S$ is supported by the source document $D$. We first generate the set of self-validated question-answer pairs from the summary, $\mathcal{P}(S)$. We then attempt to answer these questions using the source document $D$ as context. The precision score is calculated as the average semantic similarity between the summary-derived answers and the source-derived answers:

$$\text{Prec}(D, S) = \frac{1}{|\mathcal{P}(S)|} \sum_{(q,r) \in \mathcal{P}(S)} \text{sim}(\text{QA}(D, q), r) \quad (4)$$

A high precision score indicates that the claims made in the summary (represented by $r$) are consistent with the information retrievable from the source document.

### 3.6 Recall: Coverage of Important Information

Recall measures how much of the key information from the source document $D$ is preserved in the summary $S$. We generate the set of question-answer pairs from the source, $\mathcal{P}(D)$. Following QuestEval, instead of strictly comparing answer strings, we measure the *answerability* of each source question given the summary. We define an answerability score $\text{Ans}(S, q)$ based on the model's confidence in generating an answer versus producing an unanswerable token. Let $\hat{a}$ be the answer generated by the model for question $q$ given summary $S$, and let $\epsilon$ be a representative unanswerable string (e.g., "unanswerable"). We compute the length-normalized log-probability for a sequence $y$ as:

$$\ell(y) = \frac{1}{|y|} \sum_{t=1}^{|y|} \log P(y_t \mid S, q, y_{<t}) \quad (5)$$

| Source | Summary | Human Score | BanglaSummEval |
|---|---|---|---|
| মানুষের সুন্দর মুখ দেখে আনন্দিত হয়ো না। স্বভাবে সে সুন্দর নয়, দেখতে সুন্দর হলেও তার স্বভাব, তার স্পর্শ, তার রীতিনীতিকে মানুষ ঘৃণা করে। দুঃস্বভাবের মানুষ মানুষের হৃদয়ে জ্বালা ও বেদনা দেয়। তার সুন্দর মুখে মানুষ তৃপ্তি পায় না। অবোধ লোকেরা মানুষের রূপ দেখে মুগ্ধ হয় এবং তার ফল ভোগ করে। যার স্বভাব মন্দ, সে নিজেও দুশ্চিন্তাশীল, মিথ্যাবাদী, দুর্মতিকে ঘৃণা করে। মানুষ নিজে স্বভাবে সুন্দর না হলেও সে স্বভাবের সৌন্দর্যকে ভালোবাসে। স্বভাব গঠনে কঠিন পরিশ্রম ও সাধনা চাই, নইলে শয়তানকে পরাজিত করা সম্ভব নয়। | বাহ্যিক সৌন্দর্য নয়, স্বভাবের সৌন্দর্যই মানুষকে বিচারের মাপকাঠি। খারাপ স্বভাবের মানুষও বাহ্যিক সৌন্দর্যের অধিকারী হতে পারে। আর যারা খারাপ স্বভাবের তারাও সুন্দর স্বভাবের মানুষকে পছন্দ করে। তাই কঠোর পরিশ্রম ও সাধনার মাধ্যমে সুন্দর স্বভাবের অধিকারী হতে হবে। | 0.73 | 0.76 |
| আমার ১০ . ০৯ . ১৯ তারিখ থেকে ডেঙ্গু জ্বর হইছে । আজকে ১০ দিন হল । এনএস১ পজিটিভ আসছে । এর পর থেকে প্লাটিলেট কমে যাচ্ছিল । গত ৩ দিন যাবত প্লাটিলেট বাড়ছে । সর্বশেষ আজকের রিপোর্টে ১৫০০০০ আসছে । আমার শরীর কিছুটা দুর্বল । এছাড়া আর তেমন কোন সমস্যা নেই । আমি কখন বুঝতে পারব যে আমার ডেঙ্গু জ্বর ভাল হয়ে গেছে । আমার কি আর সিভিসি টেস্ট করার দরকার আছে ? | আমার ১০ . ০৯ . ১৯ তারিখ থেকে ডেঙ্গু জ্বর , আজকে ১০ দিন হল । এনএস ১ পজিটিভ আসছে । প্লাটিলেট কমে যাচ্ছিল , ৩ দিন যাবত প্লাটিলেট বাড়ছে । আজকের রিপোর্টে ১৫০০০০ আসছে । শরীর কিছুটা দুর্বল । | 0.77 | 0.77 |

Table 3: Example evaluation samples from BanglaSummEval (see Appendix B for English translation)

The answerability score is then defined as the normalized probability of the candidate answer relative to the unanswerable candidate:

$$\text{Ans}(S, q) = \frac{\exp(\ell(\hat{a}))}{\exp(\ell(\hat{a})) + \exp(\ell(\epsilon))} \quad (6)$$

To account for the relative importance of different facts, we employ a query weighting function $W(q, D) \in [0, 1]$, which predicts the importance of a question $q$ given the source $D$ (more details in the next section). The weighted recall is computed as:

$$\text{Rec}(D, S) = \frac{\sum_{(q,r) \in \mathcal{P}(D)} W(q, D) \cdot \text{Ans}(S, q)}{\sum_{(q,r) \in \mathcal{P}(D)} W(q, D)} \quad (7)$$

### 3.7 Query Weighter: Importance Weighted Evaluation

The NER module, responsible for generating candidate answers, extracts named entities and nouns but not all of these named entities carry important facts about the source $D$. Thus, questions generated from them do not contribute equally to the factual evaluation of a summary. Some questions capture central facts or entities critical to the document's meaning (e.g., "Who performed the action?" or "What was the main outcome?"), while others correspond to peripheral details. To account for this variance in informational importance, we used a *query weighter* component. The predicted query weights are used to compute the weighted recall score in Eq. 7.

The weighting mechanism is implemented using the same unified Qwen3-14B-bnb-4bit model through a lightweight scoring prompt (see Appendix A.1.4). This model was selected based on preliminary human evaluation, which demonstrated its ability to produce reliable and consistent weighting judgments. Reusing the same model across components also helps keep the overall framework lightweight and avoids introducing additional model dependencies.

Given a question and its source context, the model is asked to numerically rate the question's importance on a continuous scale from 0.0 (trivial or irrelevant) to 1.0 (highly central). The prompt explicitly distinguishes between high-importance questions which contain core entities, events, or relationships as well as low-importance ones related to minor or absent details. The final recall computation therefore becomes a weighted aggregation where each question contributes proportionally to its estimated informativeness. This ensures that factual consistency evaluation remains sensitive to the relative importance of information units rather than treating all factual elements uniformly.

### 3.8 Final Metric

The final BanglaSummEval score is the harmonic mean of the precision and recall scores, providing a balanced measure of factual consistency and important content coverage:

$$\text{Score}(D, S) = \frac{2 \cdot \text{Prec}(D, S) \cdot \text{Rec}(D, S)}{\text{Prec}(D, S) + \text{Rec}(D, S)} \quad (8)$$

## 4 Results Analysis

### 4.1 Computational Efficiency

We evaluate BanglaSummEval's computational cost on a single NVIDIA T4 GPU (16GB VRAM) available on Kaggle's or Google Colab's free tier. Table 4 reports average execution times and memory (VRAM) usage across different document lengths.

| Document Length | Time (Min) | Memory (GB) | QA Pairs |
| --- | --- | --- | --- |
| Short (∼10 entities) | 10 | 12.7 | 8–12 |
| Medium (∼20 entities) | 25 | 12.7 | 16–22 |
| Long (∼30+ entities) | 55 | 12.7 | 28–35 |

Table 4: Execution time and memory usage for BanglaSummEval on varying document lengths.

The evaluation pipeline processes an average document-summary pair in approximately 25 minutes, with linear scaling based on the number of extracted entities. The 4-bit quantization keeps peak memory (VRAM) under 13GB, enabling deployment on free-tier cloud GPUs.

### 4.2 Correlation Analysis

We validated BanglaSummEval against expert annotations on randomly sampled a total of 300 source-summary pairs both from NCTB dataset (Chowdhury et al., 2021) and BanglaCHQ-Summ (Khan et al., 2023) (see Table 3). To ensure our evaluation is not biased toward Qwen3-generated content, we deliberately selected summaries from diverse sources: the NCTB dataset, which were written by professional human writers and edited for curriculum relevance, and from BanglaCHQ-Summ, which were created by six medical informatics experts (four experts in medical informatics and two experts in both medical informatics and medicine) following standardized annotation guidelines. All 300 summaries in our evaluation set are human-written which eliminates any potential evaluator bias that could arise from preferential scoring of Qwen3-generated content. Three native Bangla speakers with 14+ years of experience rated each summary's factual consistency on a 0–1 scale. Table 5 shows strong correlation across all metrics. Pearson's $r$, which measures linear relationship between BanglaSummEval scores and human judgements shows a value of 0.694, indicating a moderately strong positive correlation between human judgement and our framework whilst the extremely low p-value ($3.84 \times 10^{-11}$) confirms that the correlation is highly statistically significant.

Spearman's $\rho$ which measures the monotonic relationship using rank-based correlation shows a value of 0.763, suggesting that BanglaSummEval ranks summaries in a very similar order to human evaluations with the p-value showing extreme statistical significance. Kendall's $\tau$ value of 0.644 also indicates strong agreement between our framework and human judgement. The coefficient of determination $R^2$ of 0.481 indicates that BanglaSummEval scores explain 48.1% of the variance in human judgments. While not perfect, owing to the fact that human judgments inherently contain subjective variations this evaluation score is a reasonable fit. Mean Absolute Error (MAE) of 0.020 indicates that on average, BanglaSummEval scores deviate by only 2 percentage points from human scores, suggesting high practical accuracy in score prediction.

Since no prior factual consistency evaluation metric specifically designed or adapted for the Bangla language exists in the literature, we could not perform direct comparisons with equivalent baselines. This positions BanglaSummEval as the first such metric for Bangla summarization, and the reported cor-relations with human judgments serve as the primary benchmark for its effectiveness in this low-resource setting.

| Metric | Value | p-value |
| --- | --- | --- |
| Pearson's $r$ | 0.694 | $3.84 \times 10^{-11}$ |
| Spearman's $\rho$ | 0.763 | $2.54 \times 10^{-14}$ |
| Kendall's $\tau$ | 0.644 | $9.04 \times 10^{-14}$ |
| $R^2$ | 0.481 | — |
| MAE | 0.020 | — |

Table 5: Correlation analysis between `BanglaSummEval` F1 scores and human judgments (n=300).

While `BanglaSummEval` does not claim to fundamentally improve LLM hallucination detection rates compared to end-to-end evaluation approaches, it provides a critical advantage by removing the black-box abstraction inherent in holistic metrics. By decomposing the evaluation process into interpretable steps such as question and answer generation, semantic matching, and question importance weighting, our framework enables researchers to identify exactly where and why an LLM fails to detect factual inconsistencies. This step-by-step breakdown facilitates targeted diagnosis of model weaknesses, whether in entity recognition, question formulation, or answer generation, thereby supporting more informed model development and refinement.

These results validate `BanglaSummEval` as an effective automated proxy for human evaluation in Bangla summarization, particularly for applications requiring transparent, interpretable quality assessment.

## 5 Implications for Low-Resource NLP Research

This study advances linguistic fairness by enabling factual consistency evaluation for an underrepresented language. The reference-free nature of the proposed framework removes reliance on human-annotated summaries, facilitating faster and lower-cost evaluation. In addition, the framework provides diagnostic insights into LLM behavior, allowing researchers to identify specific failure modes in Bangla, such as hallucinations and stages where models struggle to preserve factual information.

Despite these advantages, instruction-tuned models such as Qwen3-14B-bnb-4bit may still inherit biases originating from high-resource training data. Future work should therefore investigate bias mitigation strategies, including adversarial evaluation and systematic feedback from native Bangla speakers.

## 6 Conclusion

This paper introduced BanglaSummEval, a reference-free framework for evaluating factual consistency in Bangla summarization. The method combines question generation, question answering, semantic comparison, and importance weighting within a single multilingual instruction-tuned language model. This design enables interpretable evaluation without relying on gold-standard reference summaries. Experiments on multiple Bangla summarization datasets show strong correlation with expert human judgments. These results validate BanglaSummEval as a reliable automated proxy for factual consistency assessment.

Beyond overall scores, BanglaSummEval provides step-wise diagnostic signals. These signals help identify where factual inconsistencies occur. This supports more targeted analysis of model behavior. By addressing the lack of evaluation tools for Bangla, this work contributes to more reliable evaluation in low-resource language settings. Future work will explore domain-specific adaptation, improved semantic encoders, and extensions to other underrepresented languages.

## Limitations

There are several practical constraints that influenced our framework's design as well its performance. First, we relied on lightweight models due to limited computational resources, ensuring compatibility with freely available GPUs on platforms such as Google Colab and Kaggle. To minimize memory requirements, we used a single shared model instance (Qwen3-14B-bnb-4bit) for multiple components, e.g. question generation, question answering, and entity extraction, rather than deploying separate specialized models. To further reduce memory usage and enable deployment on limited VRAM environments, we adopted 4-bit quantization, which can lead to minor degradation in model precision.

Second, for semantic similarity scoring, we employed BERTScore using XLM-RoBERTa-base, which, while efficient, may not fully capture subtle semantic nuances in Bangla compared to larger encoders.

Lastly, the overall scoring performance is not substantially stronger than directly prompting a general-purpose LLM to assess summary quality. Our unified architecture, while computationally efficient, introduces the risk of self-reinforcing evaluation loops. Since the same model performs extraction, question generation, answering, and weighting, systematic misunderstandings of the source text may propagate through all pipeline stages, resulting in internally consistent but factually incorrect evaluations. The model may preferentially generate questions that it can confidently answer, potentially missing more discriminative questions that would better test factual consistency.

This limitation largely reflects the restricted ability of existing LLMs to understand and generate Bangla, rather than shortcomings of the framework itself. As more capable Bangla-oriented or multilingual models become available, the framework can be readily adapted without architectural modification.

These trade-offs were necessary to preserve accessibility and scalability for researchers working in low-resource settings. However, future versions of the framework could investigate mixed-precision or low-rank adaptation techniques, as well as multi-model approaches to address circular bias and better balance efficiency with performance.

While BanglaSummEval effectively detects entity errors (incorrect names, numbers, dates), it has inherent limitations in capturing complex relational errors. QuestEval, the framework we adapted, like most QA-based approaches is optimized for atomic fact extraction and comparison. Sophisticated relational errors such as complex causal chains, implicit role reversals, or multi-hop temporal dependencies may be missed if the NER and QG components do not explicitly identify these relationships as important facts. Future work could enhance relational error detection through explicit relation extraction modules or knowledge graph-based approaches.

## A Implementation Details

### A.1 System Prompts for Unified Model Components

We present the exact prompts used to instantiate the QA, QG, and NER components from the single Qwen3-14B-bnb-4bit backbone model. All prompts are formatted using the Qwen3 chat template with `enable_thinking=False` to disable internal reasoning tokens.

#### A.1.1 Question Answering (QA) Prompt

The QA component is prompted to generate a short, factual answer (1–2 words) given a context and question in Bangla.

> **QA System Prompt**
>
> **System:**
> প্রসঙ্গ এবং প্রশ্নের উপর ভিত্তি করে উত্তর দিন। শুধুমাত্র সংক্ষিপ্ত উত্তর দিন (এক বা দুই শব্দ)।
>
> **User:**
> প্রসঙ্গ: {context}
> প্রশ্ন: {question}

**English Translation:**

- System: Answer the question based on the context and question. Provide only a short answer (one or two words).

- User: Context: {context}
  Question: {question}

#### A.1.2 Question Generation (QG) Prompt

The QG component generates a short, contextually grounded question conditioned on a provided answer span.

> **QG System Prompt**
>
> **System:**
> প্রসঙ্গ এবং উত্তরের উপর ভিত্তি করে একটি সংক্ষিপ্ত প্রশ্ন তৈরি করুন।
> প্রশ্নটি এমন হতে হবে যে প্রসঙ্গে জিজ্ঞাসা করা হলে, উত্তর আপনার প্রদত্ত উত্তর হয়।
>
> **User:**
> প্রসঙ্গ: {context}
> উত্তর: {answer}

**English Translation:**

- System: Generate a short question based on the context and answer. The question must be such that when asked in the context, the response is the provided answer.

- User: Context: {context}
  Answer: {answer}

### A.1.3 Named Entity and Noun Extraction (NER) Prompt

The NER component extracts named entities and nouns from Bangla text and returns them as a comma-separated list.

> **NER System Prompt**
>
> **System:**
> আপনি একটি মডেল যা বাংলা ভাষায় প্রদত্ত পাঠ্য থেকে নামযুক্ত সত্তা এবং বিশেষ্য নিষ্কাশন করে এবং সেগুলি একটি অসংখ্যায়িত তালিকা হিসাবে প্রদান করে, কমা দ্বারা পৃথক (মোট নামযুক্ত সত্তা এবং বিশেষ্যের সংখ্যা উল্লেখ করার প্রয়োজন নেই)। কোন অতিরিক্ত ব্যাখ্যা ছাড়াই বাংলা ভাষায় আউটপুট ফেরত দিন।
>
> **User:**
> প্রসঙ্গ: {context}

**English Translation:**

- System: You are a model that extracts named entities and nouns from texts provided in Bangla and provides them as an unnumbered list, separated by commas (no need to mention the total number of named entities and nouns). Return the output in Bangla language without any additional explanations.

- User: Context: {context}

### A.1.4 Question Weighting (Importance Scorer) Prompt

The question weighter component scores each question's importance for a given source document on a scale from 0.0 to 1.0.

> **Question Weighter Prompt**
>
> **System:**
> তুমি একজন প্রশ্ন মূল্যায়নকারী। প্রদত্ত প্রসঙ্গ এবং প্রশ্ন বিবেচনা করে, প্রশ্নটি কতটা গুরুত্বপূর্ণ তা মূল্যায়ন করো।
> গুরুত্বপূর্ণ প্রশ্ন (উচ্চ স্কোর):
>
> - মূল তথ্য সম্পর্কে জিজ্ঞাসা করে (কে, কী, কখন, কোথায়)
> - প্রসঙ্গের কেন্দ্রীয় বিষয়বস্তু সম্পর্কে
> - প্রধান ব্যক্তি, স্থান, ঘটনা সম্পর্কে
>
> গুরুত্বহীন প্রশ্ন (নিম্ন স্কোর):
>
> - তুচ্ছ বিবরণ সম্পর্কে
> - প্রসঙ্গে উল্লেখ নেই এমন বিষয় সম্পর্কে
> - অপ্রাসঙ্গিক তথ্য
>
> শুধুমাত্র একটি সংখ্যা (0.0 থেকে 1.0) দিয়ে উত্তর দাও। কোন ব্যাখ্যা না।
>
> **User:**
> প্রসঙ্গ: {context}
> প্রশ্ন: {question}
> গুরুত্ব স্কোর (0.0-1.0):

**English Translation:**

- System: You are a question evaluator. Considering the given context and question, evaluate how important the question is.

  Important Questions (High Score):

  – Asks about key information (who, what, when, where)
  – About the central theme of the context
  – About main persons, places, events

  Unimportant Questions (Low Score):

  – About trivial details
  – About topics not mentioned in context

– Irrelevant information

Answer with only a number (0.0 to 1.0). No explanation.

- User: Context: {context}
  Question: {question}
  Importance Score (0.0-1.0):

## A.2 Hyperparameters

Table 6: Generation hyperparameters for all model components.

| Parameter | QA / QG / NER | Weighter |
|---|---|---|
| Model | Qwen3-14B-bnb-4bit | |
| Temperature | 0.0 | 0.0 |
| Sampling | Greedy | Greedy |
| Max New Tokens | 256 (QA), 150 (QG), 512 (NER) | 10 |
| Repetition Penalty | 1.1 | 1.0 |
| Max Sequence Length | 2048 | 2048 |

## A.3 Filtering Threshold

We set the semantic similarity threshold $\tau = 0.60$ for filtering question-answer pairs during the construction of $\mathcal{P}(C)$. This threshold was chosen empirically to balance precision (removing ill-formed QA pairs) and recall (retaining valid pairs with paraphrased answers).

## A.4 BERTScore Configuration

We use the `bert-score` Python library with the following configuration:

- **Model:** `xlm-roberta-base` (multilingual)

- **Language code:** `bn` (Bengali)

- **Metric:** Recall component only ($R$)

- **Device:** CUDA (GPU) when available

## B  English Translated Result Table

| Source (English) | Summary (English) | Human Score | BanglaSummEval |
| --- | --- | --- | --- |
| I did not find joy in seeing a beautiful face. In character, the person is not beautiful; even if visually attractive, people hate their behavior, touch, and manners. Ill-tempered people hurt others' hearts. One does not feel satisfaction from their handsome face. Naive people get dazzled by appearances and suffer consequences. Those with bad character also despise misdeeds and lies. Even if a person is not naturally good-looking, they love virtuous character. Building good character requires hard work and practice. | Outer beauty is not important; inner character is the true measure. People with bad character can still be outwardly attractive. Even those with bad character appreciate those with good character. Hence, one must strive and work hard to cultivate good character. | 0.73 | 0.76 |
| Since 10.09.19 I have had dengue fever, today makes 10 days. NS1 is positive. Platelets were decreasing, but for the last 3 days they are increasing. Today's report shows 150,000. My body is somewhat weak. There is no other major problem. When will I know my dengue is cured? | From 10.09.19 I have had dengue fever. Today is day 10. NS1 is positive. Previously, platelets were decreasing; in the last 3 days they have increased. Latest report shows 150,000. My body feels somewhat weak. | 0.77 | 0.77 |

Table 7: English translation of example evaluation samples.

## C  Human Evaluation Guidelines for Factual Consistency

**Objective**

Annotators evaluate whether generated summaries are factually consistent with source documents by identifying hallucinations, contradictions and unsupported claims on a 0.0–1.0 scale.

**Core Question**

*"Does the source document explicitly state or strongly imply this information?"*

**Scoring Rubric**

| Score | Level | Description |
| --- | --- | --- |
| 0.8 - 1.0 | Perfect | All information fully supported; no hallucinations, contradictions, or misleading inferences |
| 0.5 - 0.8 | Good | Main facts correct; minor inaccuracies or subtle unsupported details that do not alter meaning |
| 0.2 - 0.5 | Mixed | Equal mix of supported and unsupported claims; approximately half hallucinated or contradictory |
| 0 - 0.2 | Poor | Mostly unsupported or contradictory; invented details, quotes, or events |

**Evaluation Procedure**

1. **Read the source document** carefully to identify core facts, entities (names, dates, places), and narrative.

2. **Analyze the summary sentence-by-sentence** rather than holistically.

3. **Verify each claim** against the source:
   - Found in source → Correct
   - Not found in source → Hallucination penalty
   - Differs from source → Contradiction penalty

4. **Detect entity errors** (incorrect names, numbers, dates, or relationships between entities).

5. **Assign score** based on proportion of accurate vs. inaccurate claims.

**Error Categories and Penalties**

**Intrinsic Hallucination:** Direct contradiction (e.g., "black" vs. "white") — *Heavy penalty*

**Extrinsic Hallucination:** Unsupported additions (e.g., adding reasons not in source) — *Moderate-heavy penalty*

**Entity Error:** Incorrect names, numbers, or dates — *Moderate penalty*

**Relation Error:** Wrong relationships between entities — *Heavy penalty*

**Special Cases**

- **Simplification vs. Distortion:** Correct simplification of complex ideas scores 1.0; oversimplification that changes meaning scores 0.9.

- **Omissions:** Missing information affects recall, not consistency; only penalize if omission makes remaining information misleading.